# Fractal-Based Detection of Microcalcification Clusters in Digital Mammograms

## P.Shanmugavadivu[a]a*, V.Sivakumar[b]

*[a,b]Department of Computer Science and Applications,*
*Gandhigram Rural Institute – Deemed University,*
*Gandhigram – 624 302, Tamil Nadu, India*

**Abstract**

In this paper, a novel method for edge detection of microcalcification clusters in mammogram images is presented using the concept of Fractal Dimension and Hurst co-efficient that enables to locate the microcalcifications in the mammograms. This technique detects the edges accurately than the ones obtained by the conventional Sobel method. Generally, Sobel method detects the edges of the regions/objects in an image using the Fudge factor that assumes its value as 0.5, by default. In this proposed technique, the Fudge factor is suitably replaced with Hurst Co-efficient, which is computed as the difference of Fractal dimension and the topological dimension of a given input image. These two dimensions are image-dependent, and hence the respective Hurst co-efficient too varies with respect to images. Hence, the image-dependent Hurst co-efficient based Sobel method is proved to produce better results than the Fudge factor based Sobel method. The results of the proposed method substantiate the merit of the proposed technique.

## 1. Introduction

Medical Image Processing has gained momentum due to the exponential growth of applications of Digital Image Processing. Image segmentation is an important element of Digital Image Processing that subdivides the image into discrete regions/objects, each identified by the property of homogeneity of pixels, which are further subjected to high-level image processing, such as image analysis and image registration. This technique also provides the advantage of discarding unwanted image details, which are deemed as insignificant in the subsequent stages of image processing. Edge detection is a subprocess of image segmentation using which the prospective pixels that constitute contour/boundary are identified [1].

Breast cancer is quite prevalent among women across the globe. In United States, the statistics on the victims of breast cancer is quite alarming as nearly one out of every three cancers. The early detection and diagnosis of breast cancer are the vital to decrease the mortal rate by providing appropriate timely treatment [2] [3]. Microcalcifications are tiny calcium deposits that appear as bright spots in an inhomogeneous background and describe the structure of breast tissues. As mammogram images possess self-similar structure, which is the basic property of fractal objects, fractal-based techniques find wider application in almost every level of image processing.

Review of literature reveals the fact that the fractal approach is broadly used in detection and classification of microcalcifications in mammograms [4]-[6]. Fractal objects are characterized by their Fractal dimension. The Fractal dimension is a statistical quantity that explains how the fractal appears as it is magnified to finer and finer scales. Hurst Coefficient is the difference between the topological dimension and fractal dimension. In this paper, the potential of Fractal Dimension and Hurst co-efficient is combined to construct the edges that enclave the microcalcifications present in the mammograms.

————————
* Corresponding author. Tel.: +91-9443736780;
*E-mail address*: psvadivu67@gmail.com.

In this paper, section 2 describes basics of edge detection, fractal dimension and its Hurst co-efficient, section 3 discusses the computational methodology of the proposed edge detection technique paper. Section 4 deals with results and discussion and the conclusions are drawn in section 5.

## 2. Basics

### 2.1 Edge Detection

Edge detection technique uses the abrupt intensity changes among the pixels of an image. Edge is a set of connected pixels that lie on the boundary between two regions. For this purpose, a profile is defined perpendicularly to the edge direction and the results are interpreted. The magnitude of the first derivative is used to detect an edge and the sign of the second derivative can determine whether an edge pixel is on the dark or light side of an edge. The first-order derivatives in an image are computed using the Gradient operator and the Second-order derivatives are obtained using the Laplacian.

Sobel method is a proven powerful edge detection technique that computes the gradient of an image *f*, as:

$$G_x = (z_7 + 2z_8 + z_9) - (z_1 + 2z_2 + z_3)$$
$$G_y = (z_3 + 2z_6 + z_9) - (z_1 + 2z_4 + z_7)$$

(1)

where $G_x$ and $G_y$ are partial derivatives of *f* with respect to *x* and *y* respectively.

Normally, the mask for estimating partial derivative is anti-symmetric with respect to the orthogonal axis. The Sobel mask for computing x-derivative is anti-symmetric with respect to the y-axis. It has the positive sign on the right side and negative sign on the left side. The sum of all coefficients is equal to zero to make sure that the response of a constant intensity area is zero [1].

### 2.2 Fractal Dimension

Geometric primitives that are self-similar and irregular in nature are termed as fractals. *Fractal Geometry* was introduced to the world of research in 1982 by Mandelbrot. Fractals are of rough geometric shapes which can be subdivided in parts, each of which is reduced to similar of the whole [7]-[12].

Fractal objects are characterized by their Fractal dimension (defined as D).The fractal dimension is an important characteristic of fractals because it has got information about their geometric structure. The topological dimension (defined as *d*) of an object would not change whatever be the transformation an object undergoes. In the fractal world, the Fractal dimension of an object need not be an integer number and is normally greater than its topological dimension (i.e. D　d) [7] [8].

In Euclidean n-space, the bounded set X is said to be self-similar when X is the union of $N_r$ distinct non-overlapping copies of itself, each of which is similar to X scaled down by a ratio r. Fractal Dimension D of X can be derived from the relation [7], as

$$D = \frac{\log\ (N_r)}{\log\ (\frac{1}{r})}$$

(2)

### 2.3 Hurst Coefficient

The assumption of statistical self-affinity implies a linear relationship between fractal dimension, a measure of roughness and Hurst coefficient, a measure of long-memory dependence. Hurst Coefficient is defined as the difference between the topological dimension and fractal dimension. The Hurst coefficient, H is the only one parameter of interest in the fractal Brownian motion, which can be described as texture features, when we apply it to classify the breast tumors if any. Considering the topological dimension $T_d$ and fractal dimension D, the Hurst coefficient H can be calculated [2] [7] as

$$H = T_d - D \qquad\qquad (3)$$

*2.4 Morphological Operators*

The basic morphological operations used in the proposed paper are dilation and erosion. In Digital morphology, a small pattern or shape, which is known as structuring element, probes the image. For the dilation operation, the area around a pixel is set as the structuring element and the original object is allowed to grow larger. Erosion is an operation on the image in which the pixels matching the structuring element are deleted. The definitions of these operations are dependent on the image types, such as binary, gray level or color of the image being processed.

## 3. Methodology

In the proposed work, for a given input mammogram image initially the fractal dimension, D is found by box counting method using Eqn.(2). Then preprocessing is done on the input image. The morphological operations dilation and erosion are applied on the input mammogram image.

As the fractal dimension D is found, it is possible to find the Hurst coefficient H using Eqn.(3). Now, using Hurst coefficient as one of the factors along with thresholding value in the Sobel edge detection, gradient mask is obtained. To smoothen the image, post processing events dilation and erosion are applied again and the interior gaps are filled, which leads to the edge detection in an image. Finally from the segmented image microcalcification clusters are detected by outlining the border of the region(s) containing microcalcifications. The algorithm for the above methodology is given as follows.

      **Algorithm :** Detection of Microcalcification clusters from a mammogram image
      **Aim**     : To detect microcalcification clusters
      **Input**    : A 2-Dimensional mammogram image, I
      **Output** : Outlined edges of microcalcification in I

          Step 1: Read a 2-Dimensional input mammogram image I
          Step 2: [M, N]    SIZE [I]
          Step 3: If M > N then *r*    M
                Else       *r*    N
          Step 4: Compute fractal dimension D using Eqn.(2).
          Step 5: Dilate and erode I.
          Step 6: Compute Hurst coefficient H using Eqn.(3).
          Step 7: Detect the edges of I using Sobel operator with H
          Step 8: Dilate the edge detected image.
          Step 9: Fill the interior gaps in the edge detected image.
          Step 10: Smoothen the image by erosion.
          Step 11: Outline the segmented image to detect the clusters.
          Step 12: Stop.

After finding the fractal dimension D for the input mammogram image (I), morphological operations dilation and erosion were applied on I. Those image preprocessing methods were used to enhance some image features that are important for further processing. Images are inherent of randomness. As the fractal analysis is sensitive to noise, application of morphological operations tend to suppress noise if any, in addition to image enhancement on the input image.

Moreover, for comparative study, the edge detection using Sobel is done, using Fudge factor in addition to Hurst coefficient factor. Both the results are analysed. The edges constructed using Hurst coefficient is far accurate and are found to be confined to the microcalcification clusters than the one obtained with Fudge factor edges. This procedure was implemented using Matlab 7.8.

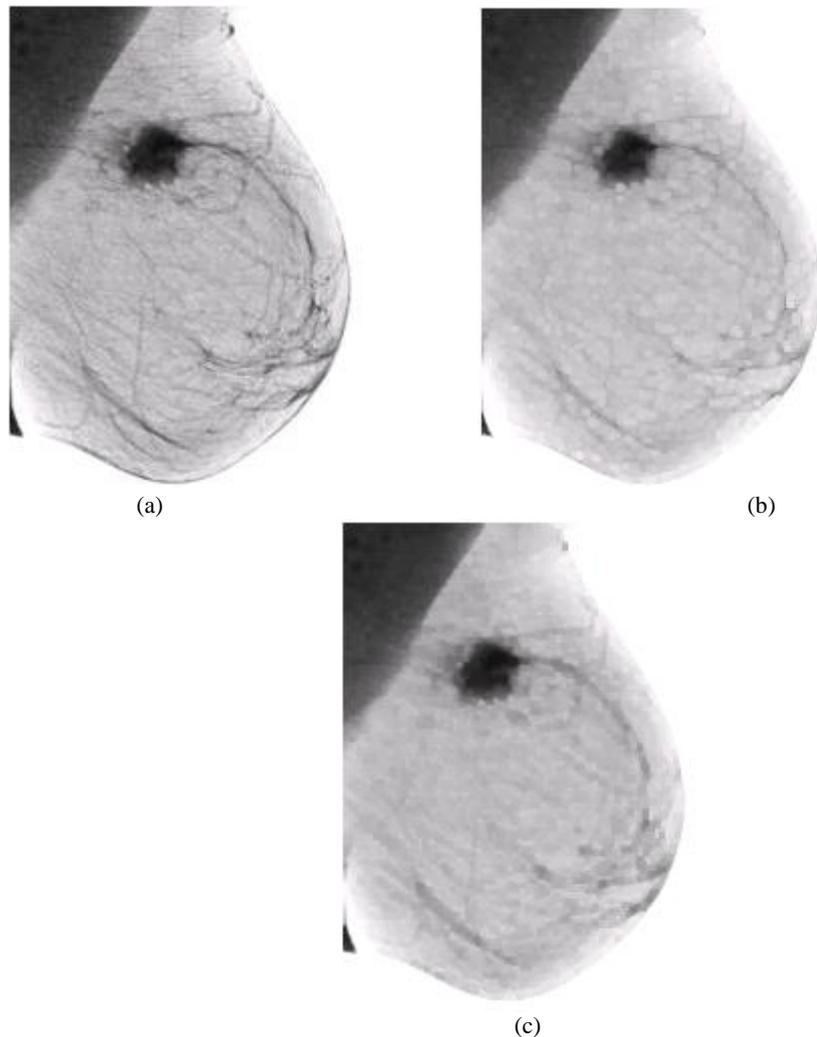

(a)                                                                              (b)

(c)

**Fig 1**: (a) Original mammogram Image (b) Dilated Image (c) Eroded Image

## 4. Results and Discussion

To assess the merit of the proposed method, different mammogram images were collected and used for the proposed study. Then as per the principle of proposed method, the edges are detected using Sobel method using Fudge factor and Hurst coefficient. The edges detected by both approaches are recorded. The proposed edge detection method is found to produce edges accurately on all those mammograms. For illustrative purpose, the results of a mammogram are depicted in Fig 1 - Fig 3.

Using Hurst coefficient in the Sobel edge detection, gradient mask is obtained. Dilation and erosion are done again as post processing method to smoothen the image so that the interior gaps are filled, that results in segmented image. Those results are shown in *Fig 2*.

Finally, by outlining the segmented portion, detection of microcalcification clusters in the image is done successfully. The detected microcalcification part is shown in F*ig 3*.

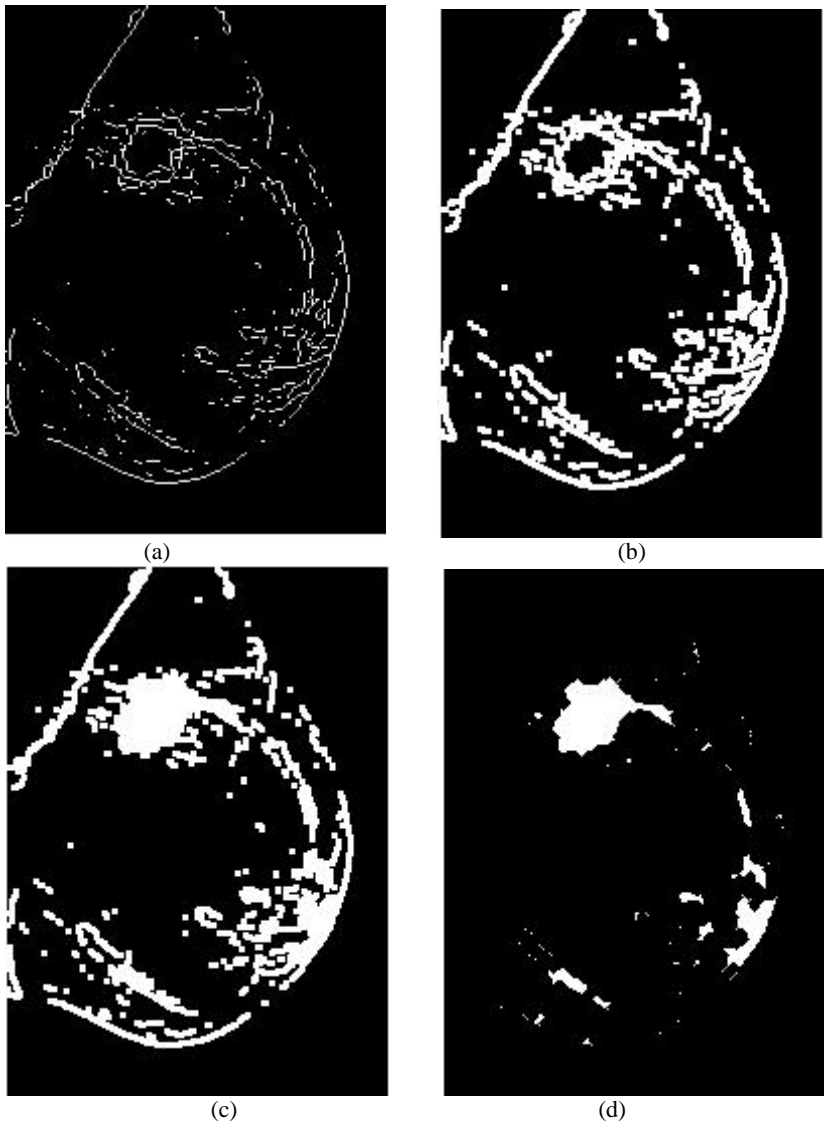

(a)                              (b)

(c)                              (d)

**Fig 2**. (a) Edges detected using Sobel operator with Hurst co-efficient  (b) Dilated image of (a)    (c) Image (b) with filled holes (d) Segmented Ouput Image.knowledgements

The microcalcifications detected by means of the proposed work i.e. by using fractal hurst coefficient is shown in fig 3. The same mass is detected by using the standard method of Sobel operator in which the fudge factor has been used as parameter factor in place of fractal Hurst coefficient. The image in which microcalcification is detected using fudge factor in standard Sobel edge detection method is shown in *Fig 4*.

Comparing the images (a) and (b) of Fig 3, it is clearly shown that the image segmentation and microcalcifications detected are far better with the result of proposed work Fig.3(b) whereas Fig.3(a) gives some unwanted additional edges. Hence it is clearly understood that the proposed work of using Fractal Hurst coefficient factor has an edge over the conventional approach.

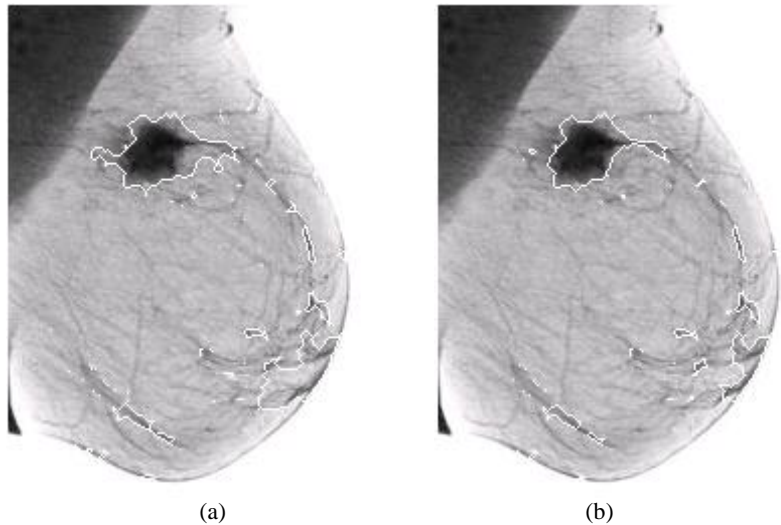

(a)                                                    (b)

**Fig 3**. Edge detection using (a) Sobel with Fudge factor (b) Sobel with Hurst co-efficient

## 5. Conclusion

This paper proposes a novel method for detection of microcalcifications in breast at an early stage using digital mammogram analysis. The proposed work brings out and highlights the merits of fractal-based Hurst co-efficient, in comparison with the Fudge factor in Sobel edge detection method. Hence, the proposed work helps in detection of microcalcifications in a mammogram image more precisely and can be used for developing an interactive expert system for an early detection of breast cancer. In the future scope of the study, after detecting and confirming the microcalcifications, classification will be done whether those microcalcifications are benign and malignant.